# NAVIGATING THE RASHOMON EFFECT: HOW PERSONALIZATION CAN HELP ADJUST INTERPRETABLE MACHINE LEARNING MODELS TO INDIVIDUAL USERS

*Completed Research Paper*

Julian Rosenberger, University of Regensburg, Regensburg, Germany, julian.rosenberger@ur.de

Philipp Schröppel, University of Ulm, Ulm, Germany, philipp.schroeppel@uni-ulm.de

Sven Kruschel, University of Regensburg, Regensburg, Germany, sven.kruschel@ur.de

Mathias Kraus, University of Regensburg, Regensburg, Germany, mathias.kraus@ur.de

Patrick Zschech, TU Dresden, Dresden, Germany, patrick.zschech@tu-dresden.de

Maximilian Förster, University of Ulm, Ulm, Germany, maximilian.foerster@uni-ulm.de

## Abstract

*The Rashomon effect describes the observation that in machine learning (ML) multiple models often achieve similar predictive performance while explaining the underlying relationships in different ways. This observation holds even for intrinsically interpretable models, such as Generalized Additive Models (GAMs), which offer users valuable insights into the model's behavior. Given the existence of multiple GAM configurations with similar predictive performance, a natural question is whether we can personalize these configurations based on users' needs for interpretability. In our study, we developed an approach to personalize models based on contextual bandits. In an online experiment with 108 users in a personalized treatment and a non-personalized control group, we found that personalization led to individualized rather than one-size-fits-all configurations. Despite these individual adjustments, the interpretability remained high across both groups, with users reporting a strong understanding of the models. Our research offers initial insights into the potential for personalizing interpretable ML.*

## 1 Introduction

As machine learning (ML) models become essential to decision-making across many industries, the need for interpretability is more critical than ever (Arrieta et al., 2020). This demand extends from high-stakes applications, such as healthcare and criminal justice, where transparency and accountability are crucial for ethical decision-making, to general business applications, where ML-driven decisions directly impact customer satisfaction, operational efficiency, and resource management. For instance, in manufacturing and service industries, ML models are used to predict demand patterns, optimize logistics, or tailor marketing strategies (Bauer et al., 2023). In these contexts, the ability to clearly interpret model predictions is essential – not only for making informed decisions that improve operational performance but also for enhancing user trust and overall experience (Ribeiro et al., 2016).

A key challenge in interpretable ML is the diversity of users and stakeholders, each with distinct preferences and cognitive styles. Thus, traditional "one-size-fits-all" interpretability approaches often face limitations in meeting the varying needs of individual users, who may differ in how they understand, trust, and act upon model explanations (Poursabzi-Sangdeh et al., 2021). These differences are also reflected in research, which paints a mixed picture of the effects of model explanations. On the one hand, two meta-studies come to the conclusion that improvements in task performance cannot be attributed to the explanation (Haag, 2024; Schemmer et al., 2022). On the other hand, positive effects can be seen when explanations are optimized for user and task (Aslan et al., 2022). Similar adaptation needs have been documented across various contexts, highlighting how human cognition and interpretation processes are highly subjective by nature (Nguyen et al., 2021; Wang & Yin, 2021). To address these challenges, a promising direction is to focus on personalization approaches that provide customized ML models tailored to users' individual needs for interpretability (Liao et al., 2020; Schröppel & Förster, 2024).

However, current research at the intersection of interpretable ML and personalization is still in its infancy, with a focus on post-hoc explanations so far. Early studies explore providing personalized explanations through adaptive conversational interfaces using generative AI and large language models (Slack et al., 2023), different types of explanations (e.g., textual, visual, mathematical) (Stokes et al., 2023; Szymanski et al., 2021), or varying configurations of post-hoc explanation methods (Aechtner et al., 2022). However, the focus on post-hoc explanations creates a fundamental tension: while these explanations can be easily adapted to different models and users, they are merely approximations of the models' decision-making processes and may not always be faithful to the underlying models (Kraus et al., 2024; Rudin, 2019).

To address this tension, we propose leveraging Generalized Additive Models (GAMs) as one class of intrinsically interpretable models, that have shown comparable predictive performance to black-box approaches on tabular data (Kraus et al., 2024; Kruschel et al., 2025; Lou et al., 2013). Our approach builds on the "Rashomon effect" in ML, which suggests that there typically exists a set of models that achieve similar, near-optimal predictive performance while explaining or modeling the underlying relationships in different ways (Breiman, 2001). This phenomenon is particularly valuable for personalization efforts, as it provides a theoretically grounded basis for offering different yet equally valid model representations to users with diverse interpretability needs. Research has demonstrated that these Rashomon sets can be systematically identified and characterized across different contexts and model classes (Fisher et al., 2019; Semenova et al., 2022), making personalization of interpretable models a practical possibility. For interpretable models like GAMs, this is particularly promising as models within this Rashomon set can exhibit substantially different visual properties in how they represent the learned patterns. For instance, GAMs based on a different number of features or with different model constraints (e.g., only allowing monotonic feature effects) frequently provide similar predictive performance. While the Rashomon effect suggests the theoretical possibility of meaningful personalization, the fundamental question remains whether users develop distinct needs for interpretability or gravitate towards similar model characteristics. We also examine how such personalization affects interpretability. Despite the recognized importance of user-centricity in interpretable ML (Brasse et al., 2023), the personalization of intrinsically interpretable models remains largely unexplored, with existing approaches primarily focusing on post-hoc explanations that may not faithfully represent model behavior (Rudin, 2019). Therefore, our research is guided by the following research questions (RQ):

**RQ1:** Does personalization of intrinsically interpretable ML models lead to individualized or one-size-fits-all models?

**RQ2:** How does the personalization of intrinsically interpretable ML models affect their interpretability?

To address these questions, we investigate the personalization of GAMs in a bike-sharing demand prediction setting, where users need to derive meaningful managerial insights. For personalization, we leverage the Rashomon effect (Breiman, 2001) and adapt an approach developed for post-hoc interpretability by Schröppel and Förster (2024). In an online experiment with 108 users split into a personalized treatment and a non-personalized control group, we found that users indeed developed distinct needs for interpretability, resulting in a diversity of individualized GAMs rather than a one-size-fits-all solution. However, this personalization did not significantly impact interpretability in terms of insight quality and user perception. Our work provides first insights into the functionality and potential benefits of personalizing intrinsically interpretable ML models.

The remainder of the paper is structured as follows. In Section 2 we describe foundations and related work. In Section 3, we outline our research approach, followed by a presentation of the results in Section 4. We conclude with a discussion of the implications of our research, a reflection on its limitations, and directions for further research in Section 5.

# 2 Foundations and Related Work

## 2.1 Interpretable Machine Learning

As ML becomes crucial for decision-making across various domains, the ability for humans to understand model decisions has become increasingly critical (Bauer et al., 2023; Janiesch et al., 2021). This understanding is crucial for users to ensure model decisions align with domain knowledge and ethical standards (Meske et al., 2022; Ribeiro et al., 2016). While current approaches primarily focus on post-hoc explanations (also known as Explainable Artificial Intelligence) that approximate model behavior after training a complex ML model, intrinsically interpretable models offer a more direct path to understanding by making their decision-making process intrinsically transparent (Arrieta et al., 2020).

For tabular data, Generalized Additive Models (GAMs) have emerged as a particularly promising class of intrinsically interpretable models. Their interpretability stems from an additive structure that can be expressed as

$$f(x) = f_1(x_1) + f_2(x_2) + f_3(x_3, x_4) + \cdots + f_n(x_n),$$

where each feature $x_i$ contributes to a shape function $f_i$ with possible pairwise interactions like $f_3(x_3, x_4)$. Modern implementations like the Explainable Boosting Machine (EBM) have enhanced this model class with ensemble learning techniques, achieving competitive performance on tabular data while remaining fully interpretable to human users (Lou et al., 2013).

The interpretability of GAMs manifests through visualizations of their learned patterns: shape plots show how individual features influence predictions, while heatmaps reveal interaction effects between features. These visualizations can vary substantially based on model configuration choices – from the number of features included to the granularity of learned patterns and the number of pair-wise interaction terms. Importantly, multiple GAM configurations can achieve similar predictive performance while differing significantly in their visual representation of the learned patterns (Bohlen et al., 2024; Kruschel et al., 2025). This phenomenon is evident in bike-sharing demand prediction applications, where GAMs effectively forecast bicycle usage, as shown in Figure 1's comparison of different GAM configurations.

The existence and diversity of multiple GAM configurations with similar predictive performance raises an intriguing question: given multiple well-performing models with different visual properties, how do we determine which configuration best serves a user's needs for interpretability?

## 2.2 Rashomon Effect

The phenomenon that multiple ML models can achieve similar, near-optimal predictive performance while having different internal structures is commonly known as the Rashomon effect (Breiman, 2001; Rudin et al., 2024). This phenomenon suggests that for many datasets, especially those influenced by noisy or uncertain factors, there may not be a single best model, but rather a set of equally valid models with diverse characteristics.

Recent research has made significant advances in identifying and characterizing these "Rashomon sets" of models. For instance, Semenova et al. (2022) developed systematic approaches to explore and map the space of high-performing models, while Fisher et al. (2019) proposed methods for efficiently identifying diverse models within the Rashomon set.

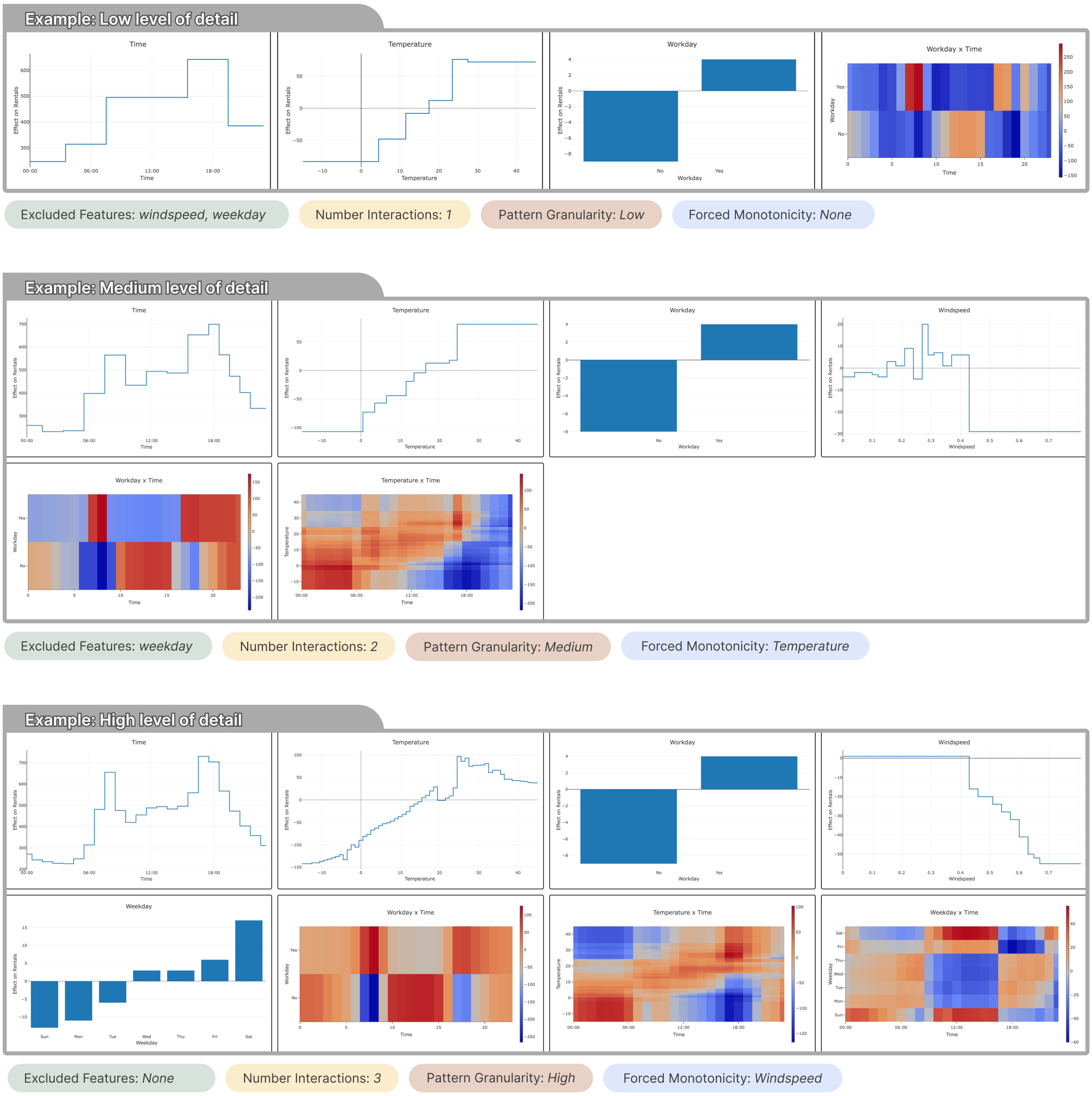

*Figure 1.* *Three (out of 92) exemplary GAM visualizations. The different complexities result from the varying hyperparameter configurations (cf. Section 3.1). At the top is a less detailed visualization with low pattern granularity, two excluded features and only one interaction. At the bottom, the other extreme, with high pattern granularity, all features and three interactions.*

For intrinsically interpretable ML models like GAMs, the Rashomon effect manifests in particularly interesting ways. Models within the Rashomon set can differ substantially in their structural characteristics while maintaining similar predictive performance. For instance, some models might achieve high performance with fewer features but more complex shape functions, while others might use more features with simpler patterns. Similarly, some models might rely more heavily on interaction effects to capture relationships, while others might achieve the same performance through main effects alone. These differences directly affect how the models visualize their learned patterns – from sparse, focused representations to more comprehensive but potentially more complex visualizations (Abdul et al., 2020).

Research has focused primarily on developing methods to identify and characterize Rashomon sets (Nevo & Ritov, 2017; Rudin et al., 2024). In the context of intrinsically interpretable ML, the question of which model from this diverse space should be presented to users remains largely unexplored. Our work addresses this gap, investigating how to leverage the rich diversity within Rashomon sets to better serve different users' needs for interpretability.

### 2.3 Personalization in Interpretable Machine Learning

Addressing distinct users' needs has been identified as a key challenge in research on interpretable ML, also referred to as "user-centricity" (Brasse et al., 2023). Research has consistently shown that interpretability is intrinsically subjective and can only be effective if it explicitly addresses users' individual needs (Hamm et al., 2023; van der Waa et al., 2021). These needs vary based on multiple factors including task context and user characteristics such as personal preferences and cognitive styles (Conati et al., 2021; Poursabzi-Sangdeh et al., 2021; Rzepka & Berger, 2018).

In response to these challenges, user-centric approaches to interpretable ML have emerged (Brasse et al., 2023). Research at the intersection of interpretable ML and user-centricity primarily focuses on adjusting post-hoc interpretability according to users' needs. To this end, researchers investigate how to design and calibrate post-hoc interpretability methods, which establish interpretability after the underlying ML model has been trained (Aechtner et al., 2022; Förster et al., 2020). Beyond, studies explore adaptive conversational interfaces (Bordt et al., 2023; Slack et al., 2023) and different types of explanations (textual, visual, mathematical) (Szymanski et al., 2021).

A notable advancement is the personalization approach by Schröppel and Förster (2024) which tailors post-hoc interpretability to individual users' needs. This translates into finding the most suitable hyperparameter configuration of a post-hoc interpretability method for each user. Building on contextual bandits, the personalization approach addresses the challenge of identifying the most effective configuration while minimizing user exposure to suboptimal interpretability, thereby balancing exploitation of known effective configurations with exploration of potentially better alternatives.

So far, research is fundamentally limited by its exclusive focus on the personalization of post-hoc interpretability. This implies that interpretability is personalized after an underlying ML model has been trained, which creates a tension between personalization and faithfulness of interpretability, as post-hoc interpretability methods remain approximations of the model's decision-making process (Kraus et al., 2024; Rudin, 2019). While studies generally suggest positive effects of personalization, the potential misalignment between interpretability and actual model behavior remains a concern (Conati et al., 2021; Schröppel & Förster, 2024). Intrinsically interpretable models address this tension by offering faithful representations of model behavior. Their interpretability is inherently linked to their structure, enabling personalization while preserving faithfulness to the model's decision-making process.

Our work addresses this gap by investigating the personalization of intrinsically interpretable models, specifically GAMs. We build on the personalization approach for post-hoc interpretability (Schröppel & Förster, 2024) and take advantage of the Rashomon effect to personalize intrinsically interpretable models. This allows us to explore personalization while maintaining predictive performance – a crucial requirement that the Rashomon effect helps satisfy. Our study constitutes an initial step towards understanding the potential of personalizing intrinsically interpretable models, which paves the way for personalized yet faithful ML interpretability.

## 3 Research Approach

### 3.1 Setting, Dataset, and Models

We embed our study in a bike-sharing demand prediction setting where participants act as managers at CityRide, a fictional bike-sharing company. This setting combines technical aspects of ML with practical business scenarios while focusing on personalization (Doshi-Velez & Kim, 2017; Ribeiro et al., 2016). The bike-sharing setting is based on a simplified version of the bike-sharing dataset (Fanaee-T & Gama, 2014) using five features ('Time', 'Temperature', 'Windspeed', 'Weekday', and 'Workday') and one year of rental records to ensure manageable interpretability in this experimental study.

The dataset is used to train different EBMs as implementations of GAMs, which we chose as our representative interpretable ML model due to their balance of predictive performance and interpretability (Kruschel et al., 2025; Lou et al., 2013). To create a diverse set of models, we define a grid of adjustable hyperparameters (cf. Table 1) with four dimensions: excluded features (4 options), number of interaction terms (3 options), pattern granularity (3 options), and forced monotonicity (4 options). The grid yields 144 possible model configurations. After filtering overlapping hyperparameter combinations (e.g., where Excluded Features = {Weekday} and Forced Monotonicity = {Weekday}), there is a total of 92 distinct hyperparameter configurations.

To yield high predictive performance, the grid is designed so that a configuration always includes highly predictive features while allowing for the exclusion of less critical ones ('Windspeed' and 'Workday'). We trained the models prior to the personalization process to evaluate the predictive performance of the GAMs trained from this set of hyperparameter configurations and to streamline the user experience during our experiment. GAMs that have been trained with one of the configurations in the grid show a comparable predictive performance and therefore lie within the Rashomon set.

| **Hyperparameter** | **Value** | **Description & Levels** |
|---|---|---|
| Excluded Features | Set | Excludes a designated set of features from the GAM.<br>{} (1), {Weekday} (2), {Windspeed} (3), {Weekday, Windspeed} (4) |
| Number Interactions | Integer | Defines the exact number of interaction terms in the GAM.<br>1 (1), 2 (2), 3 (3) |
| Pattern Granularity | Integer | Determines the maximum number of bins for main effects.<br>8 (1), 16 (2), 256 (3) |
| Forced Monotonicity | Set | Enforces the main effect of each feature in the set to be a monotonic function.<br>{} (1), {Temperature} (2), {Windspeed} (3), {Temperature, Windspeed} (4) |

*Table 1. Overview of adjustable Generalized Additive Model (GAM) hyperparameters.*

## 3.2 Personalization Approach for Intrinsically Interpretable Machine Learning

Our approach for personalizing intrinsically interpretable ML models comprises two steps: First, we adapt the personalization approach for post-hoc interpretability methods developed by Schröppel and Förster (2024) to be applicable to intrinsically interpretable models. Second, we instantiate the adapted approach to personalize GAMs in our specific experimental setting.

The adaptation of the personalization approach for post-hoc interpretability to be applicable to intrinsically interpretable models requires careful consideration of the unique relationship between model structure and interpretability. Unlike post-hoc interpretability, which can be modified independently of the underlying model, personalizing intrinsically interpretable models implies changes to the models themselves. To illustrate, enhancing a model's interpretability by reducing the number of features may come at the cost of decreased predictive performance. Selecting models exclusively from the Rashomon set allows us to optimize interpretability while maintaining predictive performance. Thus, we leverage the Rashomon effect to personalize intrinsically interpretable ML models. Accordingly, we adapt the personalization approach for post-hoc interpretability by introducing a model validation step, which ensures that users interact only with models that meet specified constraints, such as a minimum predictive performance threshold. While the characterization of the full Rashomon set is an open research problem (Rudin et al., 2022), a subset of models within the Rashomon set can be found rather easily. Our approach is based on the specification of hyperparameter sets, which leads to models within the Rashomon set. The extensive variety of near-optimal GAMs found in our specific experimental setting supports the feasibility of this approach.

We frame the personalization of intrinsically interpretable ML models as a contextual bandit problem, using Thompson Sampling to efficiently tailor models to individual users' needs. Through an iterative process (cf. Figure 2), our personalization approach systematically explores hyperparameter configurations. The process begins with the initialization of the Bayesian reward model, possibly incorporating prior information about the user and context (0). The following steps are conducted iteratively for each user:

First, based on the current probabilistic reward model, a promising hyperparameter configuration for the intrinsically interpretable ML model is selected (1). Using this selected hyperparameter configuration, the model is then trained (2) and validated against specified constraints, such as a minimum predictive performance threshold (3). If the model fails to meet the constraints, the process backtracks to step (1) to probabilistically select another hyperparameter configuration. Once a model satisfies the constraints, it is presented to the user (4). The user then interacts with the model, and the model's interpretability for the user is measured based on data gathered from this interaction. A binary reward value is computed (4) and used to update the reward model, which informs the hyperparameter selection in the next iteration (5).

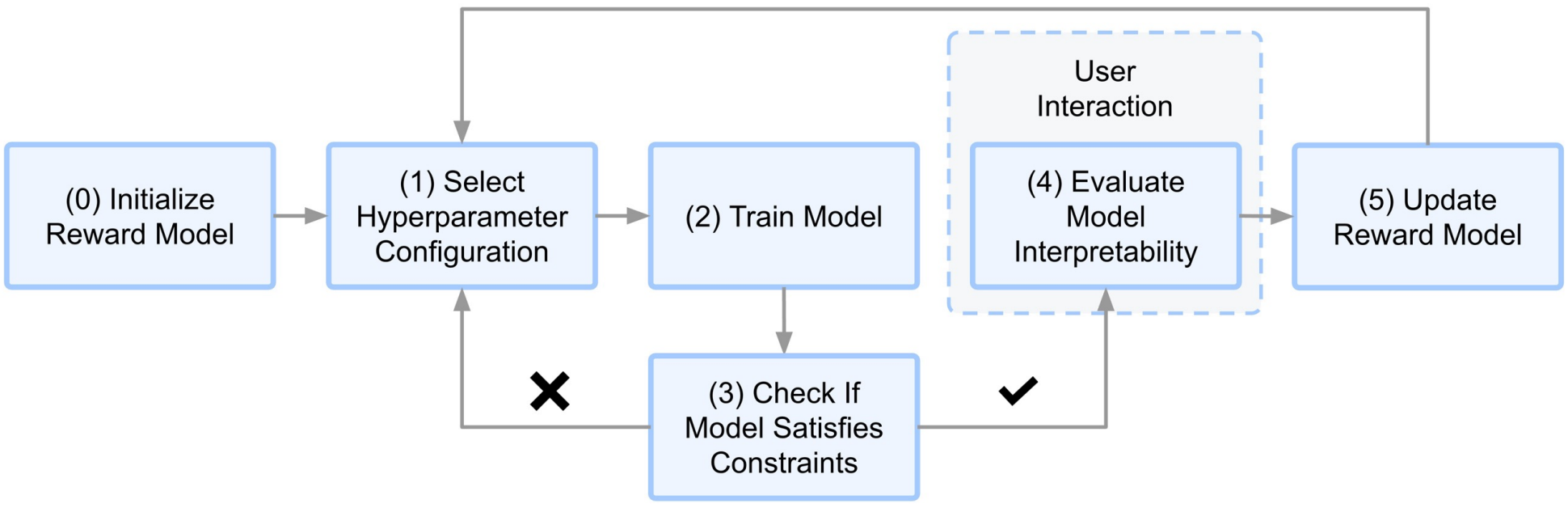

*Figure 2. Personalization approach for intrinsically interpretable ML models (adapted and extended from Schröppel and Förster (2024)).*

Instantiating the adapted personalization approach to personalize GAMs in our specific experimental setting requires three implementation choices: the selection of adjustable GAM hyperparameters, the design of a measure of models' interpretability, and the specification of a prior for the reward model's weights distribution.

First, we select a set of GAM hyperparameters to be adjusted during personalization. Our selection builds upon the hyperparameter grid evaluated during the exploration of GAMs' Rashomon set in our specific experimental setting (cf. Section 3.1) and entails four hyperparameters with a total of 92 configurations. Thus, rather than performing model validation during the personalization process, we conducted it beforehand: we trained and evaluated GAMs for each hyperparameter configuration, retaining only those configurations that produced models within the Rashomon set. This pre-validation strategy is due to experimental reasons: it provided greater control over the model selection process and enhanced the participants' experience by streamlining the experimental procedure. Importantly, this temporal shift in validation did not affect the personalization, as it merely changed when, not how, the model validation was performed. Second, we implement a mechanism to measure the interpretability of GAMs. Users rate the helpfulness of the models for generating insights using a seven-point Likert-like scale. Binary rewards are then computed from these ratings by dividing the scale into negative and positive ranges. Third, we select a prior distribution for the reward model weights, using a Normal distribution with a mean of zero and equal variance (0.5) for all weights. Both the reward scale cutoff and prior variance parameters were calibrated using data from a pre-study ($N = 30$), which also allowed us to test the entire personalization workflow and refine our measurement instruments before the main experiment.

In summary, we adapt the personalization approach for post-hoc interpretability methods to personalize intrinsically interpretable ML models (Schröppel & Förster, 2024). As a core adaption, we introduce a validation step to ensure only near-optimal performing models are considered. The Rashomon effect supports the existence of a diverse set of such high-performing models. We instantiate this novel personalization approach for intrinsically interpretable ML models in our specific setting to personalize GAMs.

## 3.3 Experimental Design and Procedure

To evaluate whether users develop distinct needs for intrinsically interpretable models (RQ1) and how personalization affects interpretability (RQ2), we designed a between-subjects online experiment with a personalized treatment and a non-personalized control group. Participants acted as resource managers at CityRide, a fictional bike-sharing company, tasked with deriving meaningful managerial insights from ML models to inform strategic decision-making. The experiment comprised three phases.

First, all participants completed an interactive tutorial using a related example. This ensured that participants could effectively interpret GAM visualizations (cf. Figure 1) before engaging with the more complex bike-sharing context. The tutorial covered the interpretation of main and interaction effects based on feature shape plots and the generation of (managerial) insights using example data to prepare participants for their role as resource managers.[1]

Second, in the personalization phase, participants explored different GAMs and rated their helpfulness for insight generation on a 7-point Likert-like scale (1 = not at all helpful, 7 = very helpful). We emphasized that their goal was to find models that would best support them in generating insights. Based on these ratings, participants in the treatment group received a GAM configuration matched to their needs through our personalization approach, while the control group received random assignments.

Finally, participants performed their task, i.e., were asked to generate five meaningful managerial insights using their assigned GAM configuration. For the treatment group, this configuration was the outcome of the personalization process, while for the control group, it was randomly selected from the predefined hyperparameter grid. The task encourages thorough exploration of the model's features and visualization characteristics, leading to a more informed final helpfulness rating. Afterwards, participants completed a post-survey to articulate their user perception.

## 3.4 Measurements and Analysis

To evaluate whether users develop distinct preferences for intrinsically interpretable ML models and how personalization affects interpretability, we employ a measurement approach that focuses on two key aspects: personalization and its impact.

First, to analyze whether personalization leads to individualized or one-size-fits-all GAMs, we examine patterns in user needs across different model configurations. Through statistical analysis of users' feedback and model selections in the personalization group, we investigate whether users develop distinct individual needs for configurations or gravitate towards similar configurations. Specifically, we analyze the reward model weights learned through Bayesian inference from users' helpfulness ratings and examine the distribution of these needs across different hyperparameter configurations.

Second, to assess the effectiveness of personalized GAMs compared to non-personalized GAMs within the Rashomon set, we evaluate both the quality of the generated insights with the help of the final GAM configuration and user perception after interaction with the final GAM configuration. For insights, we employ a structured evaluation framework with three dimensions scored from 0 to 2: specificity (from vague to highly specific), contextual integration (from no integration to significant integration), and analytical complexity (from single-variable to multi-variable interactions).

Additionally, we draw on established constructs in interpretable ML to measure user perception of interpretability (Meske et al., 2022). We included perceived usefulness and ease of use as crucial factors for successful adoption of interpretable ML (Davis, 1989), and perceived cognitive effort, which is particularly relevant for understanding users' interaction with GAMs (Abdul et al., 2020).

[1] All screens from our experimental interface, including the tutorial, are available: https://doi.org/10.17605/OSF.IO/RZB84

### 3.5 Participants

We recruited participants through the online platform Prolific. We required a minimum of 200 previous submissions and a 99% approval rate to ensure high data quality (Peer et al., 2014). Additionally, we required participants to have a minimum of 2 years of management experience to ensure that participants could realistically assume the role of resource managers in our fictional bike-sharing scenario. Despite this professional background, we focused on preparing all participants for their role through comprehensive training: an interactive tutorial using an accessible related example ensured that all participants could effectively interpret GAM visualizations before engaging with the bike-sharing context.

To incentivize thoughtful engagement with the personalization process and subsequent insight generation, participants received a base compensation of £9/hr with an additional £3 bonus opportunity based on the quality of their generated insights (Gleibs, 2017). The final sample (cf. Table 2) comprised 108 participants after excluding twelve participants: six who failed attention checks and six who exceeded the 60-minute time limit. The sample was well-balanced between the treatment group (N = 53) and the control group (N = 55), with no significant differences in gender ($\chi^2(1) = 0.002,\ p = 0.965$), management experience ($\chi^2(2) = 1.196,\ p = 0.550$), and age ($t(97) = -1.8,\ p = 0.074$).

| | | Gender (%) | | Age | Management Experience (%) | | |
|---|---|---|---|---|---|---|---|
| **Group** | **n** | **Male** | **Female** | **M ± SD** | **2-3 years** | **3-4 years** | **5+ years** |
| Control | 55 | 60.0 | 40.0 | 41.9 ± 8.8 | 10.9 | 16.4 | 72.7 |
| Treatment | 53 | 62.3 | 37.7 | 45.5 ± 11.5 | 13.2 | 9.4 | 77.4 |
| Total | 108 | 61.1 | 43.7 | 43.7 ± 10.3 | 12.0 | 13.0 | 75.0 |

*Table 2. Participant demographics and group comparisons (M = Mean, SD = Standard Deviation).*

## 4 Results

### 4.1 Convergence of Personalization

In the first step, we assess the functionality of the personalization approach. Our analysis focuses on the evolution of the reward model weights during the personalization process. The personalization approach utilizes a binary reward signal to capture user feedback on the interpretability of presented models. A reward value of +1 indicates the model's helpfulness rating was positive, while -1 indicates negative feedback. For each individual user, our personalization approach learns a reward model which estimates the expected reward. The weights of the reward model are learned via Bayesian inference and follow a Normal distribution whose mean and covariance parameters are updated based on observed reward values. Against this background, we analyze whether the reward model's weights converge and capture users' feedback.

In line with Schröppel and Förster (2024) we analyze the reward model's weights convergence through the variance of the weights distribution, which governs exploration behavior. The distribution's covariance matrices $\Sigma$ are diagonal by design. To track their evolution, we use the normalized determinant $|\Sigma|^{\frac{1}{k}}$, where $k$ represents the matrix dimensions. Higher values indicate greater variance in the associated distribution. We observe that the variance decreases over the course of personalization consistently across users (cf. Figure 3a). This pattern indicates a shift from exploration to exploitation as the system learns from users' feedback. The effect is robust, as both the 20th and 80th user quantiles show similar decreases in magnitude.

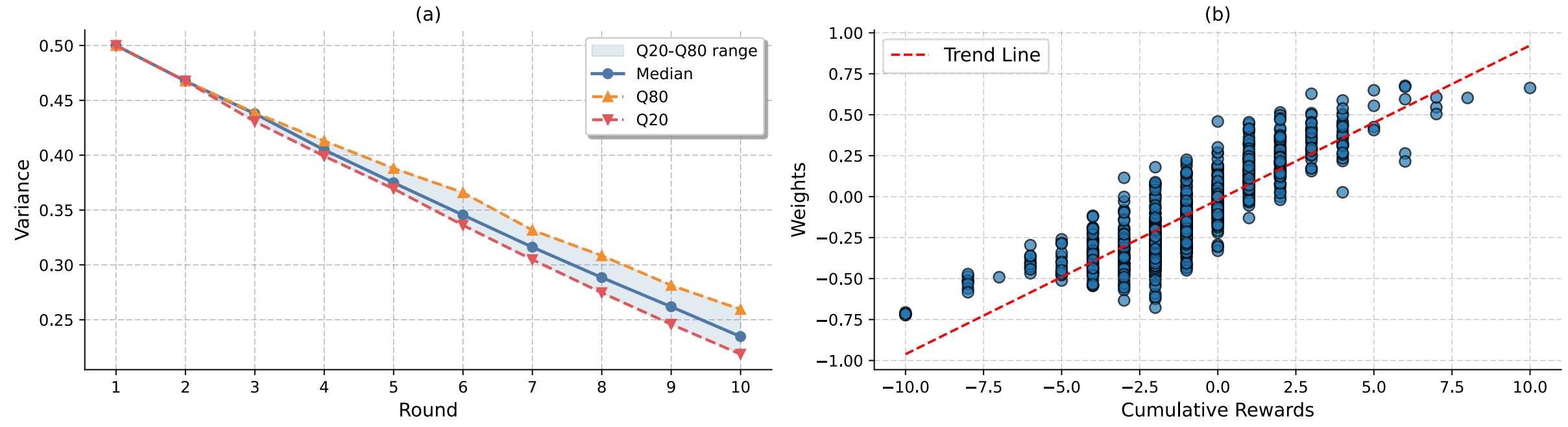

*Figure 3. Variance of the reward model's weights distribution (a) and association between cumulative reward values and mean of reward model weight distribution (b).*

To analyze whether the reward model weights capture users' feedback, we compare the cumulative reward (the sum of all reward values for models containing that hyperparameter level) with the learned mean parameter of the reward model weights. The strong association between these measures suggests that the reward model effectively captures users' feedback (cf. Figure 3b). Figure 4 demonstrates how a user's feedback translates into reward values and subsequently shapes the reward model weights, using data from a single participant in our experiment.

In summary, our results demonstrate that the reward model weights converge and effectively capture users' feedback. Yet, so far it remains unclear whether users' feedback reflect information about their needs for interpretability.

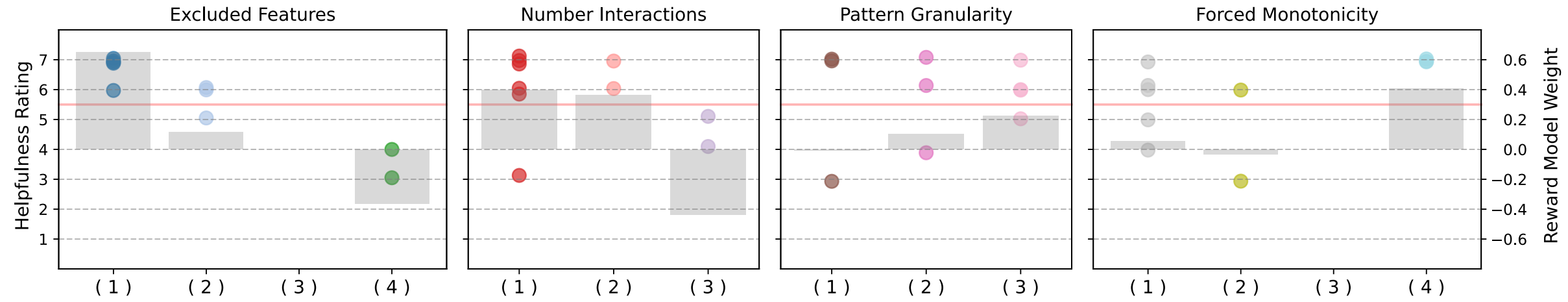

*Figure 4. Helpfulness ratings and learned reward model weights for a selected participant. The scatter plot shows user-submitted helpfulness ratings for models at this hyperparameter level. Ratings above the red line are converted to +1 reward values; those below to -1. Gray bars represent the mean of learned reward model weights.*

## 4.2 Information Gain for Different Hyperparameters

In the second step, we analyze whether the captured feedback is informative and can guide the personalization with respect to users' needs. Specifically, we assess whether users can meaningfully distinguish between models at different hyperparameter levels using the provided feedback mechanism. For a given user, we analyze the distribution of reward values for every hyperparameter level. We aim to determine whether the rewards derived from users' feedback deviate from a random coin toss. For example, for the user whose feedback is visualized in Figure 4, the ratings for hyperparameter levels Excluded Features (1, 4), Number Interactions (2, 3), and Forced Monotonicity (4) yield reward values of exclusively +1 or -1, indicating that the user was able to distinguish between models at those different hyperparameter levels. In contrast, the mixed reward values for hyperparameter levels Pattern Granularity (1), and Forced Monotonicity (1, 2) provide less informative insights into users' needs. We quantify this idea based on the concept of information gain which compares the degree of uncertainty of a random coin toss $C$ with the distribution of rewards for user $i$ and hyperparameter level $j$:

$$IG(C, R_{i,j}) := H(C) - H(R_{i,j}) = 1 - H(R_{i,j}) \in [0,1]$$

Here, $H$ denotes the Shannon entropy, a measure for a distribution's uncertainty, and $C \sim \text{Ber}\left(\frac{1}{2}\right)$ represents the distribution of a coin toss. The information gain reaches its maximum value of one when $H(R_{i,j}) = 0$, which occurs when all reward values from user $i$ for hyperparameter level $j$ are exclusively $+1$ or $-1$ (e.g., for Forced Monotonicity (1, 4), Number Interactions (2, 3), and Forced Monotonicity (4) in Figure 4). Conversely, it reaches its minimum value of 0 when reward values are evenly split between $+1$ and $-1$ (e.g., for Pattern Granularity (1, 2) in Figure 4).

Analysis of mean information gain across users and hyperparameters reveals that the collected feedback generates informative rewards that exceed random baseline performance (cf. Figure 5). Additionally, the heterogeneity in mean information gain across hyperparameters indicates different levels of feedback informativeness, suggesting that certain hyperparameter configurations elicit more informative user feedback than others. To sum up, our results indicate that the captured feedback is informative and can guide the personalization with respect to users' needs.

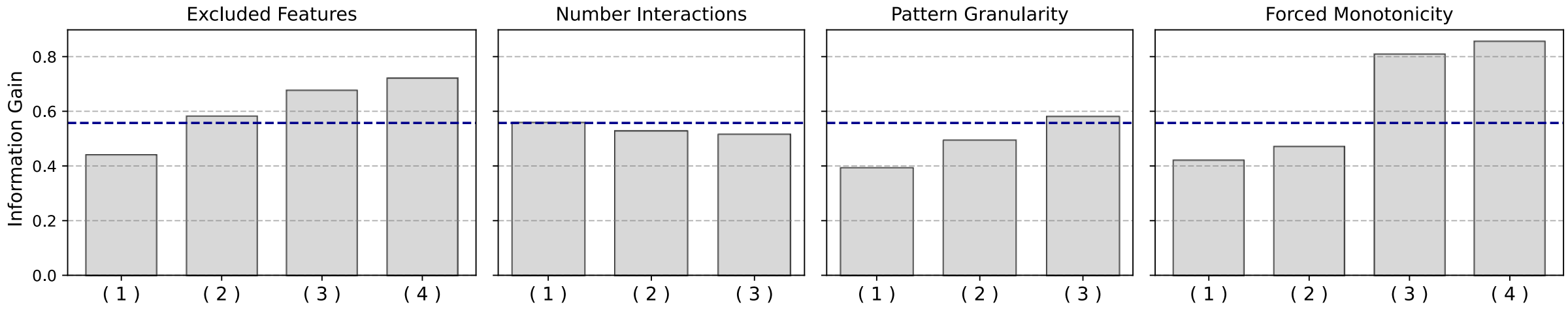

*Figure 5. Mean information gain for different hyperparameter levels. The dashed blue line indicates the mean across all information gain values.*

## 4.3 Results of the Personalization

In the third step, we examine whether personalization resulted in a single one-size-fits-all or a variety of distinct GAMs. To this end, we analyze the distribution of users' mean reward values for every hyperparameter level (cf. Figure 6). Histograms show that mean rewards are heterogeneous among users: For each hyperparameter level, some users consistently provided negative feedback while others provided positive feedback, though hyperparameters with higher information gain exhibit stronger consensus (cf. Figure 5). This consensus can take different forms – Forced Monotonicity (3) is almost universally rated negatively, suggesting it may not be useful, whereas Forced Monotonicity (4) elicits polarized but consistent opinions, indicating its value for a subset of users. This suggests that for each hyperparameter, there is not one single level suitable for all users, but different levels are suitable for different users. In conjunction with results from Section 4.1, this implies that personalization converges to different GAMs depending on a user's needs. Indeed, among the 53 participants in the treatment group, the personalization approach yielded 44 distinct final personalized GAM configurations.

## 4.4 Effects of Personalization on Insight Quality and User Perception

To address our second research question, we analyzed the impact of the final GAM configurations in the experiment (personalized in the treatment group, non-personalized in the control group) on the quality of insights generated by participants and user perception. Insight quality was evaluated using three dimensions: context integration, specificity, and complexity. Each dimension was scored on a scale of 0 to 2, with higher scores indicating better insight quality. For each insight, participants were able to score a maximum of six points. For simplicity, we averaged the scores across the five insights.

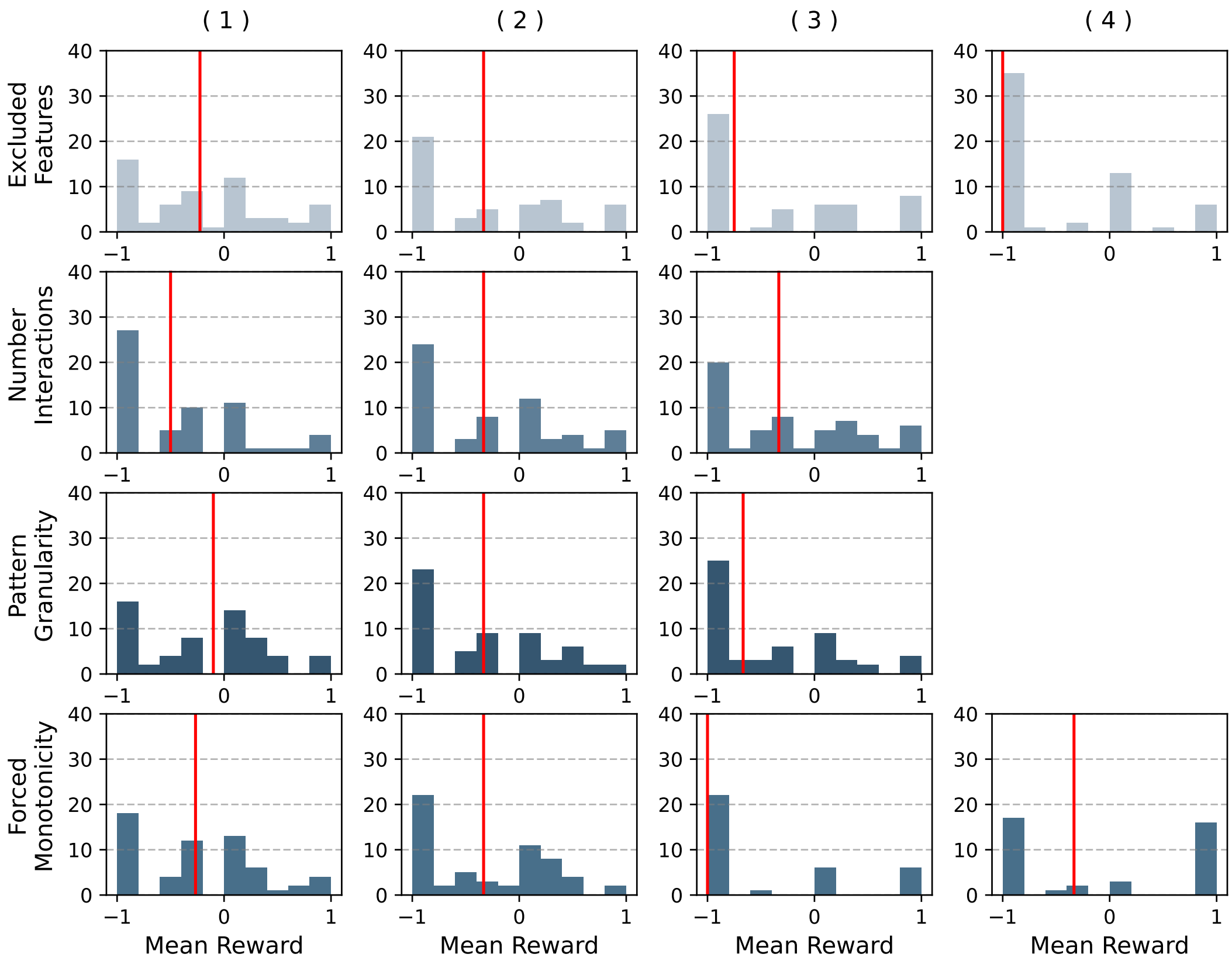

*Figure 6. Histograms of mean reward values across users. Red vertical lines indicate medians.*

Chi-square tests revealed no significant differences in the distribution of insight quality levels between the control and treatment group for specificity ($\chi^2(2) = 2.24,\ p = 0.327$) or complexity ($\chi^2(2) = 0.46,\ p = 0.794$). Although there was a statistically significant difference in the distribution of context integration levels ($\chi^2(2) = 6.31,\ p = 0.043$), Mann-Whitney U tests showed no significant differences in the ordinal levels of context integration ($W = 38686,\ p = 0.162$), specificity ($W = 37218,\ p = 0.641$), or complexity ($W = 36230,\ p = 0.895$) between the groups.

An independent samples t-test comparing the overall insight quality scores between the control group ($M = 1.88,\ SD = 1.20$) and the treatment group ($M = 1.82,\ SD = 1.24$) found no significant difference ($t(531) = 0.44,\ p = 0.659$).

In addition to insight quality, we assessed user perception, more concretely perceived helpfulness, perceived usefulness, perceived ease of use, and perceived cognitive effort. Shapiro-Wilk tests indicated that the data for all constructs were not normally distributed (all $p < 0.01$). Mann-Whitney U tests revealed no significant differences between the control and treatment groups for perceived helpfulness ($W = 1761,\ p = 0.053$), perceived usefulness ($W = 1486,\ p = 0.863$), perceived ease of use ($W = 1527,\ p = 0.673$), or perceived cognitive effort ($W = 1453,\ p = 0.978$). These findings suggest that personalization of GAMs did not significantly influence user perception compared to the non-personalized GAMs.

In summary, the personalization of GAMs did not significantly affect their interpretability, as measured by insight quality dimensions and user perception constructs.

# 5 Discussion

Our study investigates whether intrinsically interpretable models can be personalized while maintaining their key advantage of faithful representations. The results advance our understanding of interpretability personalization in several important ways.

Most importantly, we demonstrate that personalization can be extended beyond post-hoc explanations (Aechtner et al., 2022; Slack et al., 2023) to intrinsically interpretable models. This extension represents a fundamental challenge as interpretability in these models is directly tied to their structure, unlike post-hoc explanations that can be modified independently after model training (Rudin, 2019). We address this challenge by leveraging the Rashomon effect: the existence of multiple models with similar, near-optimal predictive performance but different visual representations (Breiman, 2001). By adapting a personalization approach for post-hoc explanations (Schröppel & Förster, 2024), we personalize interpretability based on 92 GAMs with comparable performance ($R^2 \geq 0.83$) that differ substantially in their visual representations (cf. Figure 1). Our empirical analysis demonstrates the practical value of this approach: users developed distinct preferences among these models, leading to 44 different configurations among 53 users in the treatment group. This diversity challenges the traditional one-size-fits-all approach to interpretability while maintaining the fidelity of intrinsic interpretability.

In our experiment, users were tasked with deriving managerial insights from GAMs in a bike-sharing context, rating the helpfulness of different model configurations for this purpose. Our analysis reveals that these helpfulness ratings provided meaningful information about users' needs regarding specific interpretability characteristics, such as the presence of interaction effects or the exclusion of features. The high information gain values across different hyperparameter levels demonstrate that users could meaningfully distinguish between different aspects of model interpretability. Interestingly, while individual users showed highly consistent preferences in their ratings (cf. Section 4.1), these preferences varied substantially between users (cf. Section 4.3). We are the first to demonstrate such distinct individual preferences for intrinsically interpretable models, suggesting that interpretability needs are more diverse than previously assumed in the literature (Poursabzi-Sangdeh et al., 2021; Ribeiro et al., 2016).

Our analysis provides initial insights into the impact of personalized interpretable ML. In our study, personalization of interpretability did not lead to significant differences in insight quality or user perception. Both groups reported high scores on additional constructs (e.g., for perceived usefulness $M = 5.98$ and $M = 5.97$ for treatment and control groups, respectively), indicating that participants could effectively work with GAMs regardless of personalization (cf. Section 3.4). The quality of insights varied similarly across both groups, ranging from detailed observations identifying "clear peaks in bike rentals around morning (8 AM - 9 AM) and evening hours (5 PM - 6 PM)" suggesting "a strong commuter pattern", to more superficial observations simply noting that rentals were "busiest on Friday and Saturday" or "increase towards the evening time". Thus, our findings suggest that either the differences in interpretability between the preferred ML models were not substantial enough to influence insight quality and user perception, or that personalization of interpretability was not a decisive factor in shaping these outcomes. In any case, our findings highlight the need for further investigation into the value of personalized interpretable ML, with our study serving as a promising starting point. Hence, we suggest the following avenues for future research:

First, since most insights simply described the plots, our approach may not fully assess participants' understanding of the underlying relationships. Second, the high similarity in insight quality and on established constructs (cf. Section 3.4) might indicate that GAMs are intrinsically easy to use or that our experimental scenario was too simplified with a limited number of features. Future studies should employ more complex datasets to determine if personalization has a greater impact in more intricate settings. Third, our participants were selected for their high knowledge and experience (cf. Section 3.5), which may have contributed to the overall high performance and limits the generalizability of our findings, as they might be able to compensate for non-optimal personalization settings due to their experience. Future research should investigate whether factors like graph literacy or domain expertise (Abdul et al., 2020) influence model preferences and interpretability effectiveness. Lastly, direct measures of interpretability that provide immediate feedback to the Bayesian reward model should be explored, moving beyond just insight generation and self-reported scales.

These findings have important implications for the development of interpretable ML. First, organizations can offer users meaningful choices by implementing our personalization approach without compromising performance. Second, we found that GAMs within the Rashomon set maintain consistent interpretability across different configurations. Even non-personalized GAMs can effectively help users to generate managerial insights.

While our study provides valuable insights, some limitations should be noted. First, our focus on GAMs in a bike-sharing context, while allowing for controlled investigation of personalization effects, may not generalize to other domains or model classes. Second, our online experimental setting, though enabling efficient collection of user preferences, might not fully capture the complexity of real-world ML model usage. Future research should extend our approach to other intrinsically interpretable models and application domains. Longitudinal studies could particularly help understand how personalization affects model usage and interpretation over time. Third, the personalization approach for intrinsically interpretable models may encounter scalability limitations, particularly when model training becomes computationally expensive due to increasing dataset and model complexity. While scalability challenges can be partially mitigated by pre-training and filtering models using a grid search approach, the adoption of more efficient methods for identifying models within the Rashomon set would enhance the feasibility of this approach in practical applications. Developing such methods represents a critical avenue for future research. Fourth, our approach to analyze the users' qualitative insights is only an initial step in investigating insight quality. In the future, we recommend incorporating inter-annotator agreement measures and more quantitative methods to strengthen the assessment of insights.

Ultimately, our work advances the field of interpretable ML by demonstrating that intrinsically interpretable models can be successfully personalized while maintaining their fundamental advantage of faithful representations. By showing that users develop distinct yet consistent preferences for different model configurations, we challenge the assumption that interpretability needs are uniform across users. This opens new possibilities for making ML models more accessible and useful to diverse users, while maintaining the crucial benefit of intrinsic interpretability: the faithful representation of model behavior.